\documentclass[conference]{IEEEtran}
\IEEEoverridecommandlockouts
\usepackage{cite}
\usepackage{amsmath,amssymb,amsfonts}
\usepackage{graphicx}
\usepackage{textcomp}
\usepackage{xcolor}
\usepackage[small,compact]{titlesec}
\titlespacing{\section}{1pt}{1pt}{1pt}
\titlespacing{\subsection}{0.5pt}{0.5pt}{0.5pt}
\titlespacing{\subsubsection}{0.5pt}{0.5pt}{0.5pt}
\usepackage[a4paper, total={184mm,239mm}]{geometry}
\def\BibTeX{{\rm B\kern-.05em{\sc i\kern-.025em b}\kern-.08em
    T\kern-.1667em\lower.7ex\hbox{E}\kern-.125emX}}
\begin{document}

\title{Class-based Quantization for Neural Networks\\
{}
} 

\author{
\IEEEauthorblockN{Wenhao Sun$^1$,
Grace Li Zhang$^2$,
Huaxi Gu$^3$,
Bing Li$^1$, 
Ulf Schlichtmann$^1$}
\IEEEauthorblockA{$^1$Chair of Electronic Design Automation, Technical University of Munich (TUM), Munich, Germany}
\IEEEauthorblockA{$^2$Hardware for Artificial Intelligence Group, TU Darmstadt, Darmstadt, Germany}
\IEEEauthorblockA{$^3$School of Telecommunications Engineering, Xidian University, Xi'an, China}
\IEEEauthorblockA{Email: \{wenhao.sun, b.li, ulf.schlichtmann\}@tum.de, grace.zhang@tu-darmstadt.de, hxgu@xidian.edu.cn}
}

\maketitle

\label{sec:abstract}

\begin{abstract}
In deep neural networks (DNNs), there are a huge number of weights and multiply-and-accumulate (MAC) operations. Accordingly, it is challenging to apply DNNs on resource-constrained platforms, e.g., mobile phones. Quantization is a method to reduce the size and the computational complexity of DNNs. Existing quantization methods either require hardware overhead to achieve a non-uniform quantization or focus on model-wise and layer-wise uniform quantization, which are not as fine-grained as filter-wise quantization. In this paper, we propose a class-based quantization method to determine the minimum number of quantization bits for each filter or neuron in DNNs individually. In the proposed method, the importance score of each filter or neuron with respect to the number of classes in the dataset is first evaluated. The larger the score is, the more important the filter or neuron is and thus the larger the number of quantization bits should be. Afterwards, a search algorithm is adopted to exploit the different importance of filters and neurons to determine the number of quantization bits of each filter or neuron. Experimental results demonstrate that the proposed method can maintain the inference accuracy with low bit-width quantization. Given the same number of quantization bits, the proposed method can also achieve a better inference accuracy than the existing methods.
\end{abstract}

% \begin{IEEEkeywords}
%     DNN, quantization, deep learning
% \end{IEEEkeywords}
\section{Introduction}\label{sec:introduction}

Deep neural networks (DNNs) have shown superb performance on tasks such as image classification\cite{Resnet} and object detection\cite{Yolov3}. However, the performance of neural networks grows along with the size. A large model such as ResNet-50\cite{Resnet} has 25.6 million parameters. Since the processors need to wait for massive weights to be loaded into the cache, the increasing number of weights not only requires more storage but also increases the inference time. These large requirements of computing and memory resources pose challenges to the deployment of DNNs on resource-constrained devices, such as mobile phones. Therefore, it is necessary to find a way to reduce the weight size of DNNs.

To address this problem, many methods, e.g., pruning and quantization, have been applied to DNNs. Pruning is an efficient way to remove weights in DNNs to reduce the size, such as \cite{pruning1, pruning2, pruning3}. In pruning a neural network, the insignificant weights are masked. Therefore, the storage requirements can be reduced, and the processors can skip the pruned weights to speed up the inference. However, pruning is a coarse-grained method, because it only has the ability to remove weights. It is difficult to decide whether the weights that are insignificant but still contribute to the accuracy of the model should be removed, which makes it hard to balance performance and efficiency. On the other hand, quantization is a fine-grained method. It quantizes the weights and activations to a low bit-width, such as 8-bits or 4-bits. Also, if weights are quantized to 0-bit, it means those weights are pruned. Therefore, besides removing useless weights, quantization can provide more flexibility to reduce the size of insignificant weights by setting the bit-width of them to a lower number. Accordingly, the model size of the neural networks and the inference accuracy can be fine-tuned to achieve a better balance compared with pruning.

There are two kinds of quantization, namely non-uniform quantization and uniform quantization. Non-uniform quantization is a method that quantizes the weights and activations with unequal quantization intervals, in which the weights and activations in the same interval share the same quantized value, such as \cite{nonuniform1, nonuniform2, nonuniform3}. For example, in ResNet-18, the distribution of weights is concentrated in the near-zero region. Hence, there should be more quantization intervals in the near-zero region to make the weights distinguishable \cite{nonuniform3}. However, the hardware implementation of non-uniform quantization is difficult \cite{nonuniform_hardware}, since it is hard to implement arithmetic operations between values with different quantization intervals. The other kind of quantization is uniform quantization, in which the quantization intervals between quantized values are equal. Uniform quantization may introduce more quantization errors, because the quantization intervals cannot be adjusted to fit the distribution of weights. However, uniform quantization can be implemented on existing neural network processors directly or with minor hardware modifications. Therefore, uniform quantization is more practical compared with non-uniform quantization when hardware implementation is taken into account.

Many methods try to improve the performance of uniform quantized networks.
\cite{uniform1} is a model-level uniform quantization method, which uses knowledge distillation to improve the performance of quantized networks. \cite{uniform2} improves the performance of model-level uniform quantized networks performed on accumulators with low bit-width by adjusting the loss function. \cite{uniform3} uses multiple settings of batch normalization layer to endow the model-level quantized neural networks with the ability to change the quantization bit-width after training, and it also uses knowledge distillation to improve the inference accuracy. \cite{uniform4} improves the training process of model-level uniform quantized networks by gradient scaling to reduce the errors in back propagation. But these approaches still ignore the flexibility of multi-bit quantization. Multi-bit quantization is an approach which quantizes the layers or filters to different bit-widths. The important layers or filters can be arranged to higher bit-width, and the insignificant layers or filters can be arranged to lower bit-width. In this way, the size of neural networks can be reduced, while the inference accuracy can be efficiently maintained. The challenge of multi-bit quantization is how to find the bit-width for different parts of the neural network. \cite{uniform5} arranges the bit-width at layer-level by reinforcement learning. However, compared with filter-level quantization, layer-level quantization is not sufficiently fine-grained. Reinforcement learning is also difficult to search the bit-widths at filter-level, since the search space is significantly larger than the search space for layer-level quantization. \cite{nonuniform3} uses a loss-based iteration method to arrange filter-level bit-width, but it focuses on non-uniform quantization and needs multiple back propagation iterations to find the best bit-width for each filter.

In this work, we propose a class-based quantization (CQ) method to find the bit-width for each filter or neuron in uniform quantization according to the user desired average bit-width. The bit-width criteria of each filter or neuron is the number of classes to which the filter or neuron is important. Given a pre-trained model, the importance scores of each filter or neuron will be collected by one-time back propagation. Based on the importance scores, the search algorithm will find the bit-width for each filter or neuron and reduce the average bit-width below the user desired average bit-width. After refining with knowledge distillation, the models will have similar performance as the original models but with much smaller average bit-width of weights.

The contributions of this work are listed as follows:
\begin{itemize} 
    \item 
    This work proposes an efficient class-based method to find the bit-width for each filter or neuron in quantization. It calculates the importance of each filter or neuron to the classes and keeps a higher bit-width on filters or neurons with higher importance scores.
    
    \item 
    The proposed algorithm only needs one-time back propagation to collect the importance scores of each filter or neuron. In the search phase, the algorithm uses inference of validation samples, such as images, instead of back propagation. Therefore, the algorithm is efficient and easy to implement.

    \item 
    Experimental results demonstrate that the proposed class-based uniform quantization method can achieve similar inference accuracy of the original models with much lower average bit-width. Compared with the existing methods, under the same bit-width setting, this method can achieve better inference accuracy.
\end{itemize}

\section{Background and Motivation}\label{sec:motivation}

\subsection{Background}
Out of the two kinds of quantization schemes, non-uniform quantization, and uniform quantization, uniform quantization is more practical and hardware friendly. Therefore, in this work, we focus on uniform quantization to determine the number of quantization bits for weights in filters or neurons.

In uniform quantization, for a full-precision input $x$ of the quantizer, the quantizer first clips $x$ to the range of [$a$, $b$], where $a$ is the lower bound of input $x$, and $b$ is the upper bound. For weights, $a$ is equal to $-b$, and the upper bound $b$ is the maximum absolute value of weights in the layer. Since ReLU is used as the activation function, the activations should always be positive. Therefore, for activations, $a$ is equal to 0. 
\begin{figure}[htbp]
    \centerline{\includegraphics[scale=0.45]{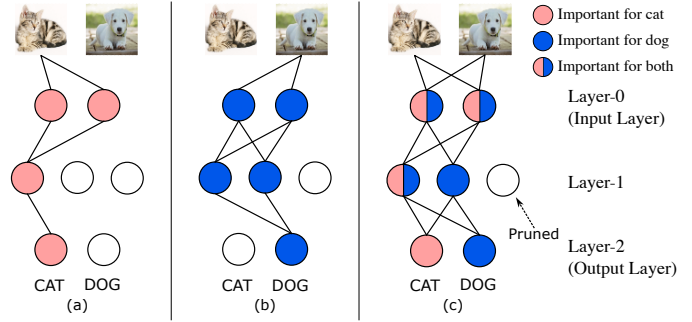}}
    \caption{An example of the data path and the importance of neurons for different classes. (a) paths for the class of cats; (b) paths for the class of dogs. (c) overlapping of the class paths.}
    \label{path}
\end{figure}
The upper bound $b$ of activations is acquired by performing inference, and it is still the maximum absolute value of activations in the layer during the inference.

The clipped value $x_c$ is defined as
\begin{equation}
    \label{eq:clip}
    x_c= \begin{cases}
        b & x \geq b  \\
        x & a < x < b \\
        a & x \leq a
        \end{cases} .
\end{equation}
Then, the clipped input $x_c$ is normalized and quantized to $x_r$ by $N$ levels, which is given by
\begin{equation}
    \label{eq:quantizer}
    x_r=\operatorname{round}\left(\left(N-1\right) * \frac{x_c - a}{b-a} \right)*\frac{1}{N-1}.
\end{equation}
Afterwards, the quantized result $x_q$ will be given by a rescaling of $x_r$:
\begin{equation}
    \label{eq:dequantizer}
    x_q= (b-a)*x_r + a.
\end{equation}

\subsection{Motivation}
The drawback of uniform quantization is that it may degrade the accuracy of the quantized neural network. To find a better way to mitigate the degradation, we propose a class-based method, where the \textit{class} is a group of images or other kinds of data sharing a same label. The concept is that different neurons have different contributions to the final outputs of the neural network, and the contribution may vary in different classes. Figure \ref{path} provides an example of this concept. It shows a multilayer perceptron (MLP), which predicts the pictures of cats and dogs. The neurons which significantly contribute to cats and dogs are not the same. Some neurons contribute only to one of the classes of cats or dogs, while some neurons contribute to both classes. The rightmost neuron in Layer-1 contributes to none of the class, so that it can be pruned.

In quantization, we assume that the neurons which contribute to many classes are more important than the neurons that contribute to fewer classes. Based on this assumption, every filter or neuron can be given an importance score, which indicates the number of classes that the filter or neuron contributes to. Then, we can use the importance score as a criterion to search the bit-width arrangement, which is the set of the quantization bits for each filter or neuron.
\section{Approach}\label{sec:method}

In this section, we will introduce the proposed class-based quantization method in detail. The goal of the quantization is to reduce the average bit-width of weights to the desired bit-width $B$ for the neural network. The quantization starts from the pre-trained full-precision model. After performing one-time back propagation, we can obtain the importance scores of each filter or neuron. Then, the search algorithm will find the bit-width for each filter or neuron. Finally, the model is quantized according to the bit-width arrangement and refined to recover the accuracy.
In the following, we describe how to calculate the importance scores of neurons in Section \ref{subsec:neurons}, and how to calculate the importance scores of filters in Section \ref{subsec:filters}. Then, we introduce the search algorithm for finding the bit-width arrangement in Section \ref{subsec:searching}. Finally, we describe the refining of the quantized neural networks in Section \ref{subsec:refine}.

\subsection{Class-based importance scores for neurons}\label{subsec:neurons}

To efficiently obtain the importance scores, we use a class-based method. In this method, the importance score of each neuron for all classes is calculated. Then, the importance score of each filter is the max score of all neurons related to the filter.

The calculation of the importance scores of each neuron for each class is based on the critical pathway theory \cite{datapath1}. As shown in Figure \ref{path}, neurons may have different contributions for different classes. 
A neuron in the critical pathway means that if it is removed, the output of the model will be changed significantly. In other words, the neuron in the critical pathway contributes significantly to the output of the neural network. Therefore, we can measure the difference of the output for an input image $x_m$ to obtain the importance score of this image. $m \in \{1,...,M\}$ is the index of class, and $M$ is the number of classes. The definition of the importance score of a neuron with respect to image $x_m$ can be written as
\begin{equation}
    \label{eq:score}
    s_{(i,j)}^{m}=\left|\Phi_{\theta}(x_m)-\Phi_{\theta}\left(x_m ; a_{j}^{i} \leftarrow 0\right)\right|
\end{equation}
where $s_{(i,j)}^{m}$ is the importance score of the neuron $j \in \{1, ..., N_i\}$ in the layer $i \in \{1, ..., L\}$ for a single image $x_m$. $L$ is the number of all layers. $\Phi_{\theta}(x_m)$ denotes the output of the neural network for sample $x_m$, and $a_{j}^{i}$ is the activation of neuron $i$ in the layer $j$. $a_{j}^{i} \leftarrow 0$ in (\ref{eq:score}) means that the activation of the neuron $j$ in the layer $i$ is frozen at zero, so that it does not participate in the computation.

The computation in (\ref{eq:score}) is intuitive, but it is very time-consuming, because we need to perform the forward propagation for $L*N_i$ times to calculate the importance scores for all neurons. To reduce the complexity, we follow \cite{datapath3} to approximately calculate (\ref{eq:score}) by Taylor expansion, which is given by
\begin{equation}
    \label{eq:scoreapprox}
    s_{(i,j)}^{m}=\left|a_{j}^{i} \nabla_{a_{j}^{i}} \Phi_{\theta}(x_m)\right|
\end{equation}
where $\nabla_{a_{j}^{i}}$ is the gradient of the output of the model with respect to the mask of the neuron $j$ in the layer $i$. In this way, we only need to perform the back propagation once to obtain the importance scores of all the neurons for image $x_m$. After the importance scores for all neurons are obtained, a threshold $\epsilon$ is used to decide whether the neurons are in the critical pathway. 
\begin{figure}[htbp]
    \centerline{\includegraphics[scale=0.458]{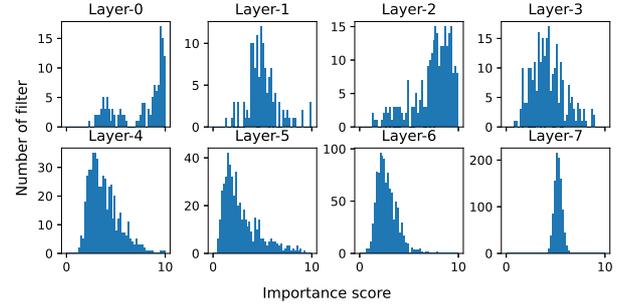}}
    \caption{Histograms of the number of filters versus the importance scores in a floating-point VGG-small\cite{vgg} network trained on CIFAR10. The x-axis shows the number of filters, and the y-axis shows the importance scores of filters.}
    \label{accbar}
\end{figure}
If $s_{(i,j)}^{m} > \epsilon$, the neuron $j$ in the layer $i$ is in the critical pathway of image $x_m$. Empirically, $\epsilon$ should be a number very close to zero. In this work, we set $\epsilon$ to $10^{-50}$.

Afterwards, a batch of validation images in class $m$ with size $N_s$ are fed to the model. By back propagation, we can obtain the set $\mathbf{s}_{j}^{i}$ including the scores of all images in the batch. 
Then, for neuron $j$ in the layer $i$, we define the importance score $\beta  _{(i,j)}^{m}$ for class $m$ as the percentage of images where the neuron is in its critical pathway, which is given by
\begin{equation}
    \label{eq:scoreoneclass}
    \beta_{(i,j)}^{m} = \frac{1}{N_s} \mid\{s \in \mathbf{s}_{j}^{i} \mid s >  \epsilon \} \mid
\end{equation}
where $s$ is the importance score of a single image for neuron $j$ in the layer $i$.
Then, the importance score of the neuron $j$ in the layer $i$ for all classes is defined as the sum of the importance scores of all classes, which is given by
\begin{equation}
    \label{eq:scoreallclass}
    \gamma_{j}^{i}=\sum_{m=1}^{M}\beta_{(i,j)}^{m} .
\end{equation}

\subsection{Class-based importance scores for filters}\label{subsec:filters}

To calculate the importance scores of each filter, we use the max score of all neurons related to the filter as the importance score of the filter to prevent ignoring the most important neurons in the filter. The definition of the importance score of a filter is given by
\begin{equation}
    \label{eq:scorechannel}
    \varphi_{k}^{i}= \max\{\gamma  \mid  \gamma \in \mathbf{\Gamma}_{k}^{i}\}
\end{equation}
where $\varphi_{k}^{i}$ is the importance score of the filter $k \in \{1,...,C_i\}$ in the layer $i$. $C_i$ is the number of filters in layer $i$. $\mathbf{\Gamma}_{k}^{i}$ is the set of importance scores defined in (\ref{eq:scoreallclass}). $\gamma$ is the importance score of a neuron in $\mathbf{\Gamma}_{k}^{i}$.

Figure \ref{accbar} shows the histograms of the number of filters versus the importance scores of the filter from a VGG-small network trained on CIFAR10. When a neuron has an importance score close to 0, it means that the neuron is not important to any class. When a neuron has an importance score close to 10 corresponding to the number of classes in CIFAR10, it means that the neuron is important to all classes. We can observe that different layers have different distributions. For example the distribution of layer-5 is skewed left, which means that most of the neurons in layer-5 are only important to a few classes. But layer-2 is skewed right and has more neurons important to more classes.

\subsection{Searching for the bit-width arrangement}\label{subsec:searching} 
\begin{figure}
    \centerline{\includegraphics[scale=0.458]{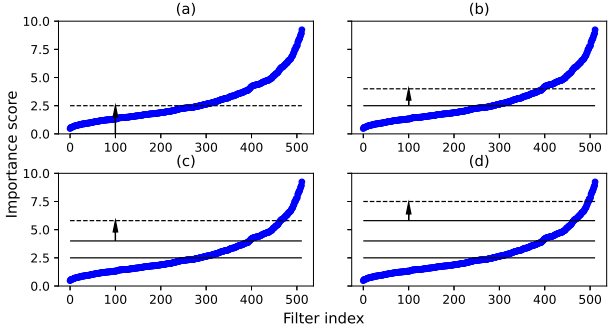}}
    \caption{An example of the search process for VGG-small \cite{vgg} on CIFAR10. The x-axis shows the indexes of the filters after sorting, and the y-axis shows the importance scores of the filters.}
    \label{searching}
\end{figure}
After the calculation of the importance score of each neuron or filter, the next step is to search for the bit-width arrangement for the quantization. The goal of this search is to reduce the current average bit-width $b_{cur}$ of the model to the desired average bit-width $B$ after the quantization of weights. As the bit-width of the model decreases, the accuracy of the model will also drop. Therefore, the challenge is how to balance the bit-widths for filters or neurons and the inference accuracy of the neural network.

Instead of directly searching for the bit-width of each filter or neuron, we first sort all the filters or neurons according to their importance scores for an efficient heuristic bit-width determination. In the example shown in Figure \ref{searching}, a curve represents the filters sorted according to the importance scores of a convolutional layer. Then, by determining some thresholds of the importance scores, the filters or neurons can be divided into several groups, where the filters or neurons in the same group share the same bit-width. Assuming that the allowed highest bit-width is $N$, we need to find $N$ thresholds, which are denoted as $p_k$, $k \in \{1,...,N\}$. For $1<k<N$, filters or neurons between the threshold $p_k$ and $p_{k-1}$ are assigned to $k-1$ bits. Filters and neurons whose importance scores are below $p_{1}$ are assigned to $0$ bits in quantization, which means that the filters or neurons are pruned. Filters and neurons whose importance scores are above $p_{N}$ are assigned to $N$ bits in quantization.

In the search process, the bit-widths of all filters and neurons are initialized to $N$. Then, the first threshold to be determined is $p_1$, which is gradually moved upward from $0$ with step $D$. As $p_1$ increases, some insignificant filters or neurons with importance scores below $p_1$ will be quantized to 0-bit and pruned, which means the inference accuracy of the neural network may start to drop. We set the target inference accuracy $T_{k}$, $k \in \{1,...,N\}$, to decide where $p_k$ should stop and be determined. $T_{1}$ is a preset value and less than the accuracy of the original neural network. For $k > 1$, the $T_{k}$ is given by
\begin{equation}
    \label{eq:target}
    T_{k} = T_{k-1} * R
\end{equation}
where $T_{k}$ and $T_{k-1}$ are the target inference accuracy of the current and previous thresholds, respectively. $R \in [0, 1]$, is a decay factor.
Once the current inference accuracy of $p_1$ is less than the target inference accuracy $T_{1}$, the threshold $p_1$ is
determined. Thereafter, for $k > 1$, the thresholds $p_k$ are determined as follows: starting from the position of $p_{k-1}$, the threshold $p_k$ is moved and the accuracy $T_k$ is evaluated similarly.
The threshold search process is repeated until all the thresholds are determined or the current average bit-width $b_{cur}$ of the neural network is less than the desired bit-width $B$. 

In case we have a very small desired bit-width $B$, after the iterations finish, the current average bit-width $b_{cur}$ may still be larger than the desired $B$. In this case, we simply move the highest bit-width threshold $p_{N}$ upward with step $D$ until reaching the maximum value of the importance scores, and the current average bit-width $b_{cur}$ is checked whether it is less than the target bit-width $B$. At this stage, changing the bit-width of filters or neurons from the highest bit-width to the second highest bit-width, such as from 4-bit to 3-bit, causes less accuracy drop than changing the bit-width of filters or neurons from the 1-bit to 0-bit, where 0-bit means that the filters or neurons are pruned. This process is repeated from $p_{N}$ to $p_1$ until $b_{cur}$ is less than $B$.

The search process is illustrated in Figure \ref{searching}. The blue curve shows the sorted importance scores of the filters in a layer of VGG-small on CIFAR10. The horizontal solid lines are the thresholds already determined, and the horizontal dashed lines are the thresholds currently searching. The target average bit-width is 2.0. We set the bit-width search range to $\{0,...,4\}$ and set $T_{1}=50\%$ and $R=0.8$. In Figure \ref{searching} (a), the threshold $p_1$ moved upward and stopped at 2.5, at that time the inference accuracy of the model is below 50\%. Then, in Figure \ref{searching} (b), the threshold $p_2$ moved upward and stopped at 4.0, at that time the inference accuracy of the model is below 40\%. The process repeats until the average bit-width reaches 2.0.

\subsection{Refining quantized neural networks}\label{subsec:refine} 
To help the model achieve a better accuracy in the refining phase, knowledge distillation \cite{KD} is applied to the full-precision model to teach the quantized model. The Loss function $L_{kd}$ in the refining phase is defined as
\begin{align}
    \label{eq:kd}
    L_{kd} &= \alpha * L_{ce} +(1-\alpha)
      \sum_{k=1}^{M} Y_k log(\frac{Y_k^{fc}}{Y_k})
\end{align}
where $\alpha$ is a factor between 0 and 1 to adjust the priority of Kullback-Leibler divergence, $L_{ce}$ is the cross-entropy loss of the original neural network. 
$\sum_{k=1}^{M} Y_k log(\frac{Y_k^{fc}}{Y_k})$ is the Kullback-Leibler divergence \cite{KL}, where $M$ is the number of classes, $Y_k^{fc}$ and $Y_k$ are the $k$-th outputs of the full-precision network and the quantized network, respectively.

In the training of the quantized neural network with knowledge distillation, it is hard to define the gradient of the quantized weights. To solve this problem, usually straight-through estimator (STE)\cite{ste} is used to update the weights in back propagation. In this work, we also use STE in the refining phase to train the quantized neural network to improve its accuracy. 
\begin{figure}[htbp]
  \centerline{\includegraphics[scale=0.458]{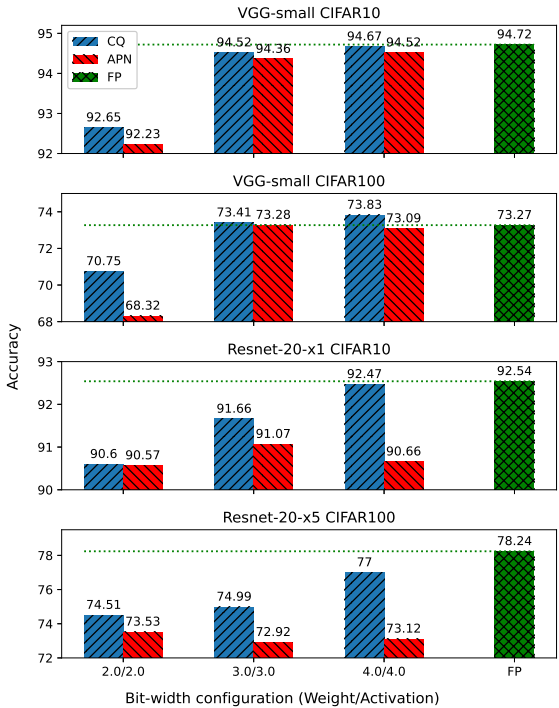}}
  \caption{Comparison of accuracy between CQ and APN \cite{uniform3} with 2.0/2.0, 3.0/3.0, and 4.0/4.0 bit-width settings and full-precision models. The blue bars are the proposed method, and the red bars are APN. The green bars are the full-precision baseline models in \cite{Resnet} and \cite{vgg}.}
  \label{baracc}
\end{figure}

\section{Experimental Results}\label{sec:results}

To demonstrate the performance of the class-based quantization (CQ), three neural network configurations, VGG-small adopted from \cite{vggsmall}, ResNet-20\cite{Resnet} with expand-1 (ResNet-20-x1) and expand-5 (ResNet-20-x5) were applied to two datasets, CIFAR10\cite{CIFAR} and CIFAR100\cite{CIFAR}, respectively. 
The algorithm and neural networks were implemented with Pytorch on Nvidia Quadro RTX 6000 GPUs. 

We compared CQ with Any-precision network (APN) \cite{uniform3} and WrapNet (WN) \cite{uniform2} under equal conditions. The results of APN were obtained using the source code provided on GitHub \cite{uniform3}, and neural networks of APN were set to individual bit-width. The results of WN were adopted from \cite{uniform2}. In the training phase, the learning rate was initialized to 0.1 for ResNets and 0.02 for VGG-small, and it was divided by 10 at 100th, 150th, and 300th epochs. The momentum was set to 0.9, and the weight decay was set to 0.0001 for ResNets and 0.0005 for VGG-small. The batch size was set to 100 for all datasets, and training was stopped after 400 epochs. 

The bit-width arrangement of weights was set according to Section \ref{sec:method}, and activations were directly set to the desired bit-widths.
In the refining phase, all the parameters of the optimizer were the same as in the training phase. In all networks, the first layer and the output layer were not quantized as in \cite{uniform2} and \cite{uniform3}. 
\begin{figure}[htbp]
  \centerline{\includegraphics[scale=0.458]{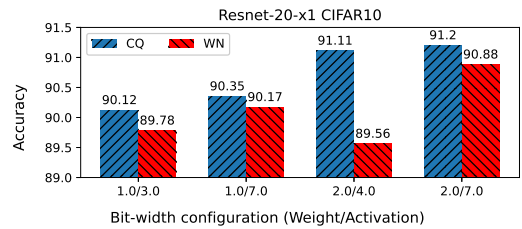}}
  \caption{Comparison of accuracy between CQ and WN \cite{uniform2} with 1.0/3.0, 1.0/7.0, 2.0/4.0, and 2.0/7.0 bit-width settings. The blue bars are the proposed method, and the red bars are WN.}
  \label{baracc2}
\end{figure}
\begin{figure}[htbp]
  \centerline{\includegraphics[scale=0.458]{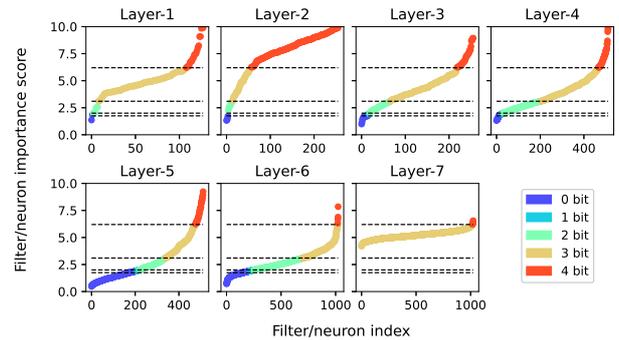}}
  \caption{Sorted filter importance score distribution of VGG-small with 2.0/2.0 bit-width on CIFAR10. The x-axis shows the indexes of the filters after sorting, and the y-axis shows the importance scores of the filters or neurons.}
  \label{layerscores}
\end{figure}
Because in CQ, the different filters or neurons may be quantized to different bit-widths, in the following experiments, the desired bit-width settings of weights are the average of all quantized weights and denoted as $\frac{\sum^N_{i=1}b_i}{N}$, where $N$ is the total number of weights except for the first layer and the output layer, and $b_i$ is the bit-width of the $i$-th weight.
The knowledge distillation loss was applied in the refining phase. $\alpha$ in (\ref{eq:kd}) was set to 0.3.

The comparison between the accuracy of CQ and APN is shown in Figure \ref{baracc}. The bit-widths are set to 2.0/2.0, 3.0/3.0, and 4.0/4.0 in the format of \textit{weight}/\textit{activation}, because the bit-width of the weights and activations in APN can only be set to the same number. The results show that CQ can achieve better accuracy than APN on every bit-width setting. In VGG-small of CIFAR10 and CIFAR100 with 3.0/3.0 and 4.0/4.0 settings, both CQ and APN are close to the full-precision model, but CQ still achieves better results. Note that VGG-small on CIFAR100 with 3.0/3.0 and 4.0/4.0 settings even outperforms the floating-point network. This is because of the regularization effect of the quantization, as pointed out in \cite{nonuniform3}. In VGG-small on CIFAR10 and CIFAR100 with 2.0/2.0 settings, CQ is better than APN for 0.42\% and 2.43\%, respectively. In ResNet-20-x1 on CIFAR10, CQ and APN are close on 2.0/2.0 setting, but CQ is better than APN on 3.0/3.0 and 4.0/4.0 settings. In ResNet-20-x5 on CIFAR100, CQ is significantly better than APN on all bit-width settings.

In figure \ref{baracc2}, it shows the accuracy comparison of ResNet-20-x1 on CIFAR10 between CQ and WN. The bit-width settings are 1.0/3.0, 1.0/7.0, 2.0/4.0, and 2.0/7.0 as in \cite{uniform2}. 
The results show that CQ can achieve better accuracy than WN on all bit-width settings. Especially in 2.0/4.0 setting, the accuracy of CQ is 1.5\% higher than WN. We can also observe that the accuracy of CQ is more stable with lower activation bit-width settings.
\begin{figure}[htbp]
  \centerline{\includegraphics[scale=0.458]{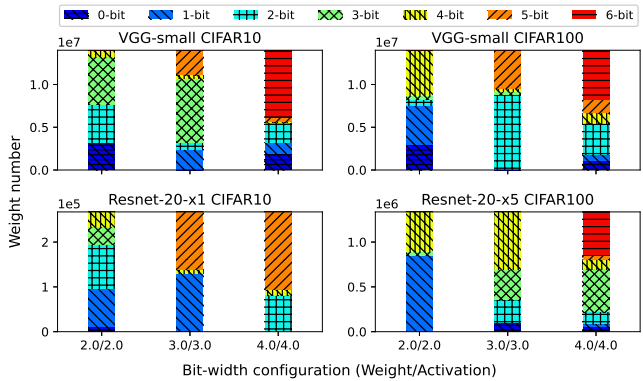}}
  \caption{Bit-width percentage of all neural networks with 2.0/2.0, 3.0/3.0 and 4.0/4.0 bit-width setting.}
  \label{bitallbar}
\end{figure}

As shown in Figure \ref{layerscores}, we take VGG-small with 2.0/2.0 setting on CIFAR10 as an example to demonstrate the bit-widths arrangement. The horizontal lines are the thresholds of the different bit-width settings. 
From bottom to top, the thresholds of 0/1-bit, 1/2-bit, 2/3-bit, and 3/4-bit are 1.9, 2.0, 3.1, and 6.2, respectively. 
The layers except for layer-2 and layer-7 have similar distributions, where considerable numbers of the filters have lower importance scores, meaning they only contribute to images from a few classes and should be quantized to lower bit-width. Especially for layer-5 and layer-6, which are the fully-connected layers, many neurons have been quantized to 0-bit. Layer-1, layer-3, and layer-4 have smaller percentage of filters lower than 1-bit. Instead, they have more filters in 2-bit and 3-bit, which indicates that these layers have more insignificant filters, but these filters still contribute to the outputs. 
On the contrary, layer-2 has more filters with higher scores. They are important for almost all images and should be quantized to high bit-width. 
The layer-7, which is the last layer before the output layer, has no filter with quantized weights lower than 2-bit, because it needs more neurons than other fully-connected layers to represent the output classes.

Figure \ref{bitallbar} shows the percentages of all models with all bit-width settings. We can see that all models have utilized the flexibility of multi-bit quantization. The VGG-small network has more filters quantized to 0-bit, most of which are in the fully-connected layers. ResNet-20-x1 and ResNet-20-x5 should keep more filters in 1 and 2 bits instead of 0 bits, because pruning in the convolutional layers can cause the bigger accuracy drop than other bit-width. In 4.0/4.0 settings, the neural networks can keep more filters in high bit-width. Therefore, they can achieve high inference accuracy which is very close to the full-precision models. In 2.0/2.0 and 3.0/3.0 settings, more filters are quantized to low bit-width to balance the filters in high bit-width. The high-precision filters contribute more to the accuracy, which allows the neural network to keep its accuracy even in low bit-width settings.

\section{Conclusion}
\label{sec:conclusion}

In this paper, we have proposed a class-based quantization scheme for DNNs, which is based on the importance scores of neurons and filters to determine the bit-widths. Experimental results demonstrated that with a small average bit-width of quantization, the inference accuracy can still be maintained with the proposed method. In addition, under the same bit-width settings, the proposed method achieved a better inference accuracy than other existing methods.

\let\oldbibliography\thebibliography
\renewcommand{\thebibliography}[1]{%
\oldbibliography{#1}%
\fontsize{8.0pt}{8.0}\selectfont
\setlength{\itemsep}{0.1pt}%
}

\bibliographystyle{IEEEtran}
\bibliography{bibfile}

\end{document}